\documentclass[letterpaper]{article} 
\usepackage[submission]{aaai23}  
\usepackage{times}  
\usepackage{helvet}  
\usepackage{courier}  
\usepackage[hyphens]{url}  
\usepackage{graphicx} 
\usepackage{natbib}  
\usepackage{caption} 
\usepackage{algorithm}
\usepackage{algorithmic}

\usepackage{newfloat}
\usepackage{listings}
\DeclareCaptionStyle{ruled}{labelfont=normalfont,labelsep=colon,strut=off} 
\floatstyle{ruled}
\newfloat{listing}{tb}{lst}{}
\floatname{listing}{Listing}
\usepackage{algorithm}
\usepackage{algorithmic}
\usepackage{multirow}
\graphicspath{{./figure/}}
\DeclareGraphicsExtensions{.pdf,.jpeg,.png,.jpg}
\usepackage{booktabs}
\usepackage{amsmath}

\title{Bridging the Emotional Semantic Gap via Multimodal Relevance Estimation}

\author {
    Chuan Zhang \textsuperscript{\rm 1},
    Daoxin Zhang \textsuperscript{\rm 1},
    Ruixiu Zhang \textsuperscript{\rm 1},
    Jiawei Li \textsuperscript{\rm 2},
    Jianke Zhu \textsuperscript{\rm 1}
}
\affiliations {
    \textsuperscript{\rm 1} Zhejiang University\\
    \textsuperscript{\rm 2} Alibaba Group\\
    \{zhangchuan, dxzhang, 22051027, jkzhu\}@zju.edu.cn, mingong.ljw@alibaba-inc.com
}

\usepackage{bibentry}

\begin{document}

\maketitle

\begin{abstract}
Human beings have rich ways of emotional expressions, including facial action, voice, and natural languages. Due to the diversity and complexity of different individuals, the emotions expressed by various modalities may be semantically irrelevant. Directly fusing information from different modalities may inevitably make the model subject to the noise from semantically irrelevant modalities. To tackle this problem, we propose a multimodal relevance estimation network to capture the relevant semantics among modalities in multimodal emotions. Specifically, we take advantage of an attention mechanism to reflect the semantic relevance weights of each modality. Moreover, we propose a relevant semantic estimation loss to weakly supervise the semantics of each modality. Furthermore, we make use of contrastive learning to optimize the similarity of category-level modality-relevant semantics across different modalities in feature space, thereby bridging the semantic gap between heterogeneous modalities. In order to better reflect the emotional state in the real interactive scenarios and perform the semantic relevance analysis, we collect a single-label discrete multimodal emotion dataset named SDME, which enables researchers to conduct multimodal semantic relevance research with large category bias. Experiments on continuous and discrete emotion datasets show that our model can effectively capture the relevant semantics, especially for the large deviations in modal semantics. The code and SDME dataset will be publicly available.
\end{abstract}

\section{1 Introduction}
Affective computing plays an important role in multimodal understanding, which is widely used in video understanding, medical treatment, safety production and other fields. Being an important part of affective computing, emotion is an efficient representation of a character's mental state. Our work aims to perform multimodal emotion analysis to understand the sentimental state of characters.

\begin{figure}[tbp]
	\centering
    \includegraphics[width=\linewidth,height=0.6\linewidth]{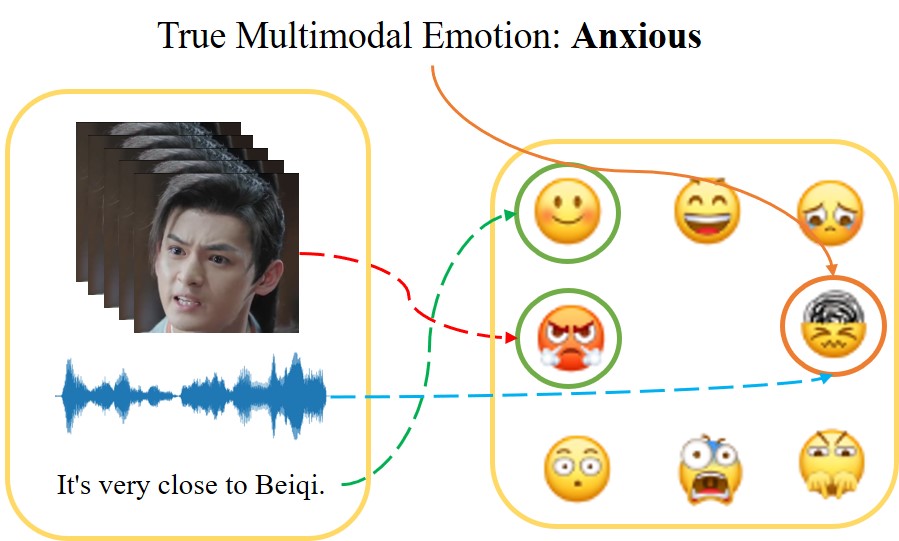}%
	\caption{An example of irrelevant multimodal emotion semantics with different visual, audio, and textual semantics.}
	\label{fig1}
	\vspace{-0.25in}
\end{figure}

Multimodal emotion analysis introduces linguistic information and nonverbal behavior, which extracts the rich semantic information from natural language, facial action, and acoustics. Compared with the model having single-modality, it is able to judge the character's emotion more accurately and robustly so as to reduce the emotional divergence. Due to the complexity of the character's affection, the emotional expression of each modality often has semantic irrelevant problems. For example, an anxious emotion may be treated as anger according to the frown feature from the visual point of view, anxiety from the audio point of view and neutral in natural language, as shown in Figure \ref{fig1}. Therefore, judging the dominant modality and learning the semantically relevant feature representation information in each modality is very important for efficient feature fusion.

In this work, we propose a multimodal fusion framework to extract the relevant semantic features of each modality. Emotion data always contains a lot of inter-modal semantic noise. Most of the existing methods do not take into account the noise among modalities in information fusion, which interferes with the effective learning of the model. To reduce the influence of noisy semantics on multimodal fusion, we try to calculate the relevance weight of each sub-modal feature with respect to the multi-modal feature and model the semantic information between emotional modalities before multimodal fusion, which predicts the sample relevance in a instance-level manner. Thus, we can weight the relevant semantics based on the attention mechanism in order to reduce the impact of noisy information. We design a relevant semantic estimation loss to perform weak supervision on the relevance weighted results. This ensures that the model can learn the effective relevant semantic information. In addition, we introduce contrastive learning to deal with relevant semantic features, which transforms vanilla noise contrastive estimation loss from instance-level to category-level. Thereby, it further bridges the semantics gap between samples of the same category.

Due to the difficulty in collecting the emotional data, the existing multimodal emotion analysis testbeds basically only have the continuous emotion (such as MOSI \cite{zadeh2016mosi}, MOSEI  \cite{zadeh2018multimodal}, ERATO \cite{gao2021pairwise}) and multi-label discrete emotion (such as MOSEI \cite{zadeh2018multimodal}, IEMOCAP \cite{busso2008iemocap}). There is a lack of independent single-label discrete emotion datasets. As continuous emotion datasets usually have only positive, neutral and negative samples, the differences between categories are fuzzy. On the other hand, the multi-label discrete emotion dataset contains a large number of complex emotions, there is no clear definition of each category. At the same time, almost all the existing emotion analysis datasets do not have unimodal annotations. Due to the complexity of human emotion, the semantics of each modality often contains different information, where the unimodality information is essential to analyzing the emotional semantic information. To facilitate the research on multimodal emotion analysis in modal semantic relevance, we propose a single label discrete multimodal emotion dataset (SDME), which has eight emotion categories with clear semantic deviation and multimodal annotations.

Our main contributions can be summarized as follows. Firstly, a multimodal relevance estimation network (MRE) is proposed to reduce the semantic gap between heterogeneous modalities by evaluating the semantic relevance of each modality. Secondly, we collect a new dataset named SDME with multimodal and unimodal annotations, which fills the vacancy of single-label discrete emotion dataset in emotion analysis research. Finally, experimental results on three datasets show that our proposed approach is effective.

\section{2 Related Work}

\subsection{2.1 Multimodal Emotion Datasets}

To facilitate the multimodal emotion analysis, various multimodal emotion datasets are collected. MOSI, MOSEI, SIMS~\cite{yu2020ch} and ERATO provide the positive, neutral and negative continuous emotion labels. In addition to providing multimodal annotation, SIMS provides unimodal annotation for the study of potential emotion. Based on the three classes, ERATO provides a more refined five-category annotation to evaluate the dialogue relationships. In addition, MOSEI and IEMOCAP contain multi-label discrete annotation data to model human emotional states. Due to the complexity of human affections, the existing continuous emotion dataset and multi-label discrete emotion dataset are vague in definition. We hope to fill the vacancy of a single-label discrete multimodal emotion dataset and clearly identify each emotion category. To this end, a single-label multimodal emotion dataset is collected for semantic relevance analysis, which provides single-modal and multimodal annotations. 

\subsection{2.2 Multimodal Emotion Analysis}
Multimodal emotion analysis is an important research topic in multimodal learning, which has already attracted lots of attention. Some studies focus on the representation of multimodality \cite{qi2021zero,ju2020transformer,liang2020semi,nie2020c,mittal2021affect2mm}. Others investigate the fusion part of multimodality \cite{khan2021exploiting,chen2021learning,shen2020memor,lv2021progressive}. Besides, a few research is concerned with the lack of modalities \cite{yuan2021transformer}. In this work, we focus on analyzing the extraction and fusing the relevant sentiment semantics in multimodality, which provides a detailed analysis of common different fusion methods.

TFN \cite{zadeh2017tensor} proposes a tensor fusion network, in which the tensor representation is obtained by calculating the outer product between unimodal representations. LMF \cite{liu2018efficient} presents a low-rank multimodal fusion method to reduce the computational complexity of tensor-based methods. MFN \cite{zadeh2018memory} designs a memory fusion network with a special attention mechanism for cross-view interaction. Graph-MFN~\cite{zadeh2018multimodal} extends MFN with a dynamic fusion graph. MULT \cite{dai2021multimodal} proposes a cross-modal transformer that enhances the target modality from another source modality by learning the attention between the two modalities. Self$\_$MM \cite{yu2021learning} uses a self-supervised strategy to focus on single modality feature modeling. MISA \cite{hazarika2020misa} aims to learn modality-invariant and modality-specific representation. Differently from the above methods, we aim to efficiently fuse the semantic model by capturing the relevant semantics among modalities. 

\subsection{2.3 Contrastive Learning}

Contrastive learning tries to represent the features by mapping the similar inputs to their adjacent points in the potential feature space, which aims to maximize the interaction information between different views or extract it between local and global features from the same sample. Contrastive learning is often used in self-supervised learning. Recently, it has been studied in the fields of knowledge distillation, image generation, and multimodal learning~\cite{chen2021distilling,morgado2021audio,jeon2021feature,chen2021transferrable,tao2020self,zhang2021enhancing}. Based on the vanilla contrastive learning loss~\cite{dyer2014notes,oord2018representation}, we construct a class-level contrastive evaluation loss for semantic noise. Semantic estimation is performed on the extracted inter-modal relevant semantics to bridge the gap of modal semantic. 

\begin{figure*}[tbp]
	\centering
    \includegraphics[width=\linewidth,height=0.38\linewidth]{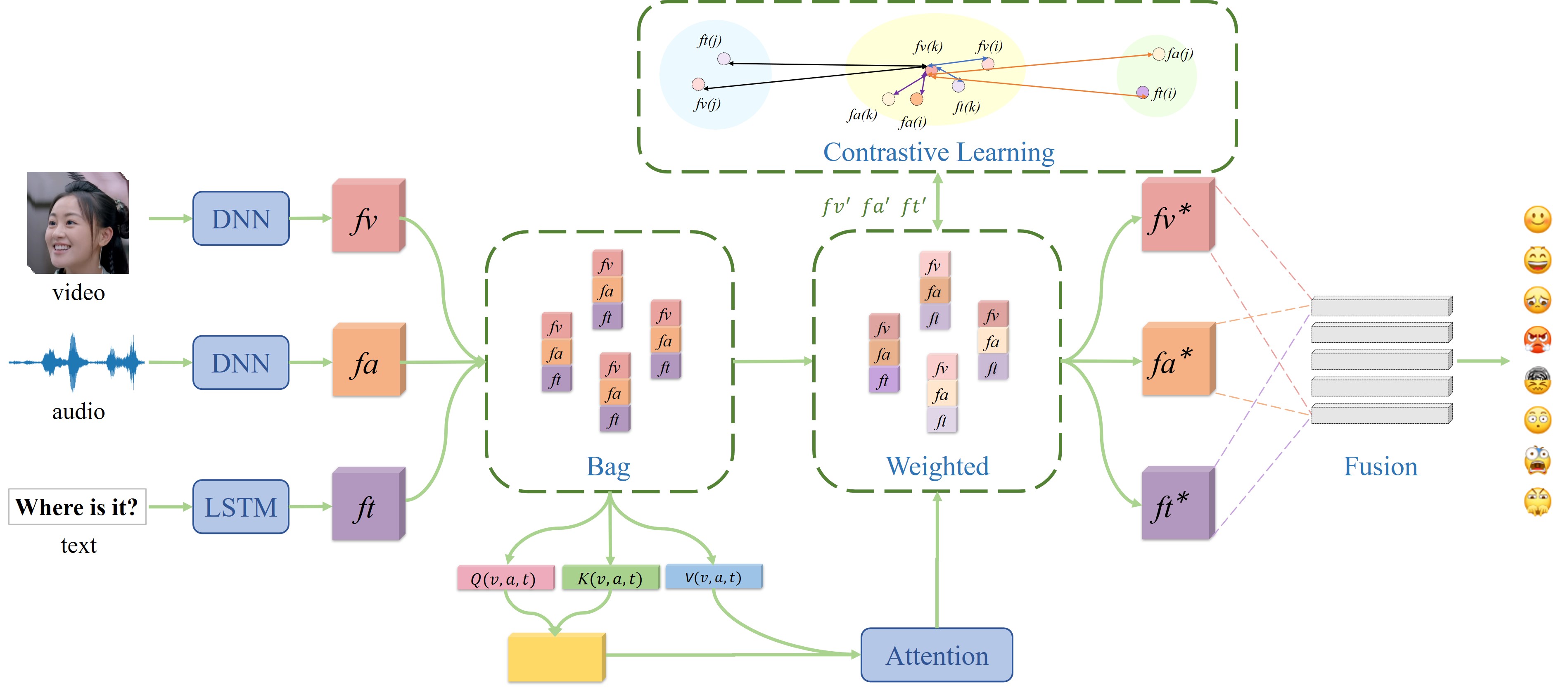}%
	\caption{The Overall Framework of Multimodal Relevant Estimation Network. The network combines the unimodal features into the instance bags, which is weighted by n relevant semantic estimate by attention mechanisms and weakly supervised methods. Afterward, the model performs feature distance optimization on the weighted category-level semantic estimation and obtains the relevant semantic representations of each modality. Finally, each unimodal feature is dimensionally reduced and fused to obtain the final prediction result.}
	\label{fig3}
	\vspace{-0.2in}
\end{figure*}

\section{3 Approach}
We propose a multimodal relevance estimation network for multimodal emotion analysis, which extracts relevant semantics between various modalities. To ensure efficient multimodal fusion, contrastive learning is employed to optimize the latent feature space. In this section, we firstly introduce the overall model architecture in Section 3.1. Secondly, we elaborate on the core relevant semantic estimation and contrastive semantic estimation modules in Sections 3.2 and 3.3, respectively. Finally, we give the final optimization objective in Section 3.4.

\subsection{3.1 Overall Architecture}
The multimodal relevance estimation network consists of four modules, including feature representation, relevant semantic estimation, contrastive semantic estimation, and low-rank fusion. We aim to learn the key semantics of multi-modalities and reduce the influence of noise features on multimodal fusion by extracting the relevant semantics between each modality. To this end, we bridge the inter-modal and intra-modal semantic gaps by the relevant semantic estimation and contrastive semantic estimation modules, respectively. The overall structure is summarized in Figure \ref{fig3}.

We extract the feature for each modality contained in the data sample. As in~\cite{yu2021learning}, we employ the off-the-shelf tools to extract the visual and audio feature representations. Specifically, OpenFace~\cite{baltrusaitis2018openface} is used to extract facial features and facial action units in the video as visual representations. Moreover, Librosa~\cite{mcfee2015librosa} is used to extract MFCC, zero-crossing rate, and Constant-Q spectrum as audio representations. By taking advantage of the language models, we make use of the pre-trained 12-layer BERT~\cite{devlin2018bert} to extract the textual features. Then, we feed the visual and audio representations into a 3-layer DNN network, and the textual representations is treated as the input of an LSTM network for feature extraction.

We feed the obtained visual, audio, and textual representations into the relevant semantic estimation module and the contrastive semantic estimation module to bridge the relevant semantics between each modality at the intra-modal and inter-modal levels. Thereby, it reduces the effect of noise features on multi-modality and extract the core semantics. In the relevant semantic estimation module, we employ an attention mechanism to get the relevant semantic weight and evaluate the contribution of each modality to multimodal semantics, thereby weighting the semantics of each modality and bridging intra-modal semantics. At the same time, we make use of the relevant semantic weight to adaptively estimate the weights of the loss function to ensure reasonable parameter updates in a weakly supervised manner. In the contrastive semantic estimation module, we focus on bridging inter-modal and inter-category relevant semantics, which further distinguishes the distribution of relevant semantic information in the feature space obtained in the relevant semantic evaluation module. The details will be explained in the following. Figure~\ref{module} illustrates our proposed module.

Due to the high dimensionality of modal features, we map the multimodal features into low-rank tensors in the phase of modal fusion, which greatly reduces the total number of parameters for fusion feature  and improves the computational efficiency. We employ the low-rank weight and the deviation tensor with modal features as the input of linear layers, which finally obtains the fusion representation for classification. Let $S_{i}^{r}$ denote the low-rank features of each modality, which can be computed as follows:
\begin{equation} \label{rank}
S_{i}^{r}=\bigotimes_{r=1}^{R} s_{r},
\end{equation}
where $\bigotimes_{r=1}^{R}$ denotes the tensor outer product over a set of vectors indexed by r, and $s_{r}$ is the input unimodal representation.

We obtain the low-rank features of each modality and fuse them by product. Thus, the final feature classification head $\mathcal{F}$ is obtained by:
\begin{equation} \label{eq5}
\mathcal{F}=W_{m}^{r} \cdot \prod_{i=\{v, a, t\}} S_{i}^{r}+\operatorname{Bias}_{m}^{r},
\end{equation}
where $W_{m}^{r}$ and $\operatorname{Bias}_{m}^{r}$ are the low-rank weight and bias.

\subsection{3.2 Relevant Semantic Estimation}

\begin{figure}[tbp]
	\centering
    \includegraphics[width=\linewidth,height=0.7\linewidth]{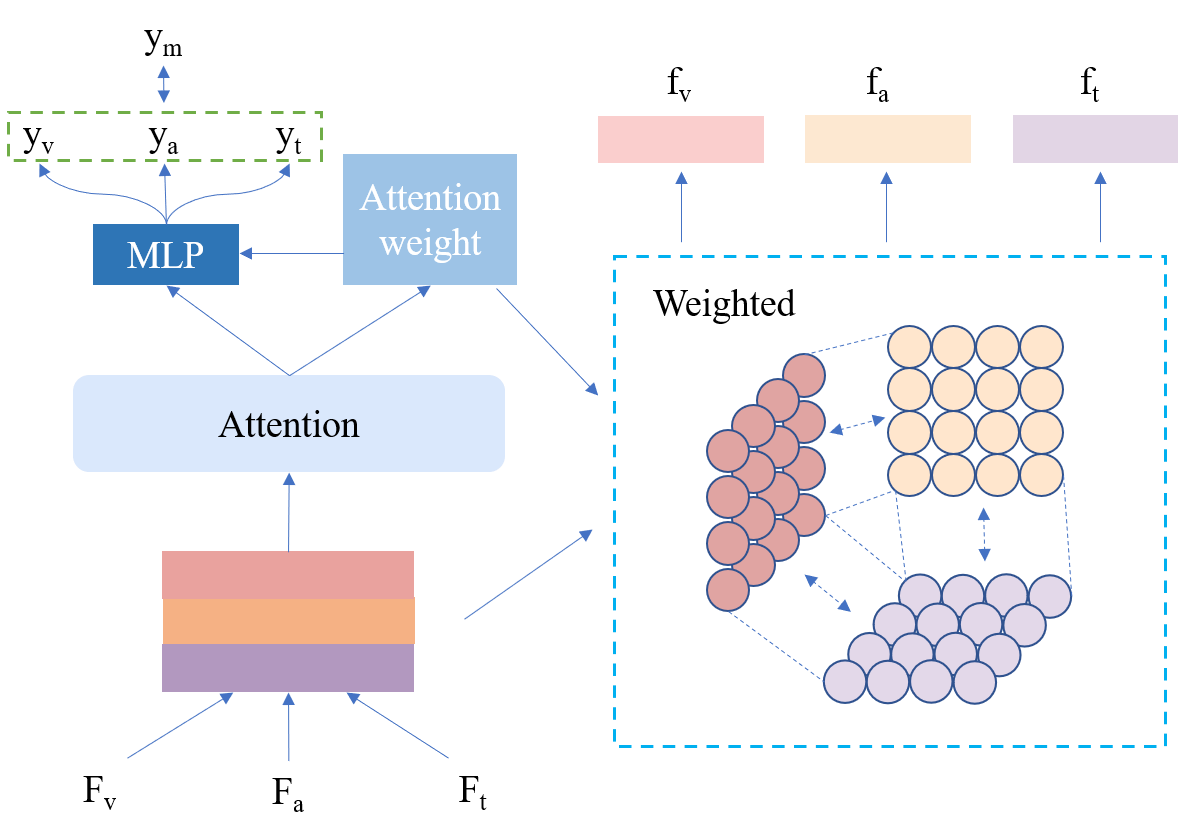}%
	\caption{The extracted unimodal features are stacked and fed to the attention module, where the modality-level relevance weight estimation and category-level relevance estimation are performed respectively.}
	\label{module}
	\vspace{-0.2in}
\end{figure}

Multimodal emotion analysis aims to determine the emotions from the visual ($I_v$), audio ($I_a$), text ($I_t$) and other information. In multimodal emotion, at least one sub-modal has the same semantics as the multimodal emotion. Therefore, it can reduce noise information to a certain extent by extracting the key relevant semantics expressing multimodal emotion in sub-modalities, which may promote the efficiency of fusion.

In multimodal tasks, each sample consists of several sub-modalities of information. Due to the complexity of multimodal data, we usually only get the multimodal annotation of a dataset. However, in some multimodal tasks such as multimodal emotion analysis, the semantics information expressed by each modality is not identical. Fusing the features of each modality directly may result in semantically irrelevant noise, which may affect the final fusion efficiency. As at least one sub-modal has the same semantics as the multimodal emotion, we can reduce noise information to a certain extent by extracting the key relevant semantics expressing multimodal emotion in sub-modalities, which may promote the efficiency of fusion. We are committed to mining the key semantics and relevant semantics between modalities by the reasonable weights for multimodal fusion. We make a self-attention on multimodal bag samples composed of multiple unimodal features to obtain the multimodal inter-modal semantic attention $A$ as below:
\begin{equation} \label{att}
\begin{aligned}
A_{\{v, a, t\}} &=\operatorname{softmax}\left(\frac{Q_{\{v, a, t\}} K_{\{v, a, t\}}^{\top}}{\sqrt{d_{k}}}\right) V_{\{v, a, t\}}.
\end{aligned}
\end{equation}
Under the self-attention mechanism~\cite{vaswani2017attention}, $Q$, $K$, and $V$ use the same input and are composed of stacks of various sub-modalities, such as vision, audio and text modalities.%

As we map the extracted modal features to the same dimension before, it is very efficient to compute the semantic attention. We feed the attention results into two MLP networks, $M_{1}$ and $M_{2}$, to predict the relevance between modalities and the relevance between classes in different modalities, respectively. We employ the instance-level features of the attention results to predict the unimodal with the highest relevance weight via $M_{1}$. Since the difference in the feature distribution of heterogeneous multimodal data help us to make an instance-level prediction, we use bag labels $Y_M$ to supervise multi-modal features as well as use them as pseudo-labels $Y_I = \{Y_V, Y_A, Y_T\}$ to perform weak supervision on the prediction results of each unimodal via $M_{2}$. Since the unimodal semantics and multimodal semantics are unnecessarily the same, we adopt the relevant semantic weight obtained by $M_{1}$ to weight the $M_{2}$ relevance loss. Therefore, the model can pay more attention to the predicted unimodal features that are closer to the multimodal core semantics, as shown in the left of Figure \ref{module}.

At this point, we employ the modal relevant weight $\mathcal{H}$ obtained from $M_{1}$ to judge the contribution of sub-modalities to multimodal samples. As emotions are consistent over short periods of time, we use the mean of sampling features to represent the features of samples. Also, the semantic weighted attention is employed to obtain inter-modal relevant semantic information from each data sample $x_{k}$. Therefore, we perform a weighted calculation on the initial modal features as below:
\begin{equation} \label{eq4}
\hat{S}=\mathcal{H} \odot\frac{1}{N} \sum_{k=1}^{N} x_{k},
\end{equation}%
where $N$ is the sampling amount of modal features and $\hat{S}$ is the semantically weighted embedding of the stacked individual modalities. $\hat{S}$ may effectively represent the relevant semantics between modalities, where sub-modalities should have similar feature distributions. 

To learn from the relevant semantic estimation, we design a relevant semantic loss $L_{rs}$ as follows:
\begin{equation} \label{eq6}
\begin{aligned}
L_{rs} &=-\sum_{s=\{v, a, t\}} \sum_{i=1}^{N} (1+\mathcal{H}_{i}^{\top}) \log P_{s}\left(Y_{s} \mid \hat{\mathcal{H}}_{i} ; \theta_{rs}^{s}\right) \\
&-\sum_{i=1}^{N} \log P_{m}\left(Y_{m} \mid \hat{\mathcal{H}}_{i} ; \theta_{rs}^{m}\right).
\end{aligned}
\end{equation}%

$\mathcal{H}_{i}$ is the relevance weight of a single modality, which is used to make the model emphasize the loss of relevant modality. $\hat{\mathcal{H}}_{i}$ is the category weight get from $M_{2}$. $Y_{m}$ is the multimodal label and $Y_{s}$ is the pseudo-label for each sub-modal. $P_{s}$ and $P_{m}$ are the predicted probabilities of the sub-modals and multimodal features, respectively. It employs weak supervision to enable the model to get more effective intra-modal relevant semantic features. We use multimodal labels to weakly supervise the semantically weighted unimodal semantic features, hoping the obtained modal relevance semantic features can well represent the multimodal core meaning. We judge the contribution of each modality to the final fused multimodal semantics according to the semantically relevant attention obtained before, so as to assign reasonable weight to the loss results of each modality. This ensures that the model can learn the key semantic information more effectively.

\subsection{3.3 Contrastive Semantic Estimation}

To further bridge the multimodal semantic gap and ensure the validity of the semantic information among sub-modalities, we introduce contrastive semantic estimation to optimize the semantic information relevant to the same samples while obtaining a highly efficient semantic association.

We employ contrastive learning to optimize the semantic feature relevance between samples. According to the principle of contrastive learning, we intend to reduce the feature distance between positive samples and push negative samples away in order to optimize the modal representation in the latent feature space. To this end, we optimize the noise contrastive estimation loss based on infoNCE~\cite{oord2018representation}:
\begin{equation} \label{eq7}
P_{nce}(x_{1}, x_{2})=\frac{\exp \left(\Phi\left(x_{1(i)}, x_{2(i)}\right) / \tau\right)}{\left.\sum_{j=1}^{n} \exp \left(\Phi\left(x_{1(j)}, x_{2(j)}\right)\right) / \tau\right)},
\end{equation}%
where the temperature $\tau$ is a smoothness parameter. InfoNCE treats each example pair as a separate individual while ignoring the feature similarity of similar samples to a certain extent. Given that the relevant semantics of the same kind of emotion is similar, they can be recombined without changing the core semantics. 

We modify the infoNCE loss to perform contrastive learning in units of classes respectively, so as to bridge the distance between the relevant semantics of the contract samples in the latent feature space. Considering each individual as a class will misjudge some positive samples as negative samples, we reduce the penalty for contrastive learning of positive samples and increases the penalty for negative samples. We design a contrastive semantic estimation loss as follows:
\begin{equation} \label{eq8}
\resizebox{.91\linewidth}{!}{$
    \displaystyle
L_{c n c e}=\sum_{s_{1},s_{2}\in\{v, a, t\}}\left(-\frac{1}{N_p} \sum_{i=1}^{N_p} \log P_{i}^{N_p}(s_{1},s_{2})-\frac{1}{N_n} \sum_{i=1}^{N_n} \log(1-P_{i}^{N_n}(s1,s2))\right)
$},
\end{equation}%
where $s_{1}$ and $s_{2}$ are the two different sub-modals. $N_{p}$ is the total number of positive samples in the same class, and $N_{n}$ is the number of negative samples. 

\subsection{3.4 Learning Objective}

We employ the cross-entropy loss to classify the multimodal emotional data. Moreover, $L_{sa}$ is used to bridge the intra-modal and inter-modal features. $L_{cnce}$ penalizes the potential relevance of features within and between modalities. The overall loss of our multimodal multi-instance contrastive estimation network can be formulated as below:%
\begin{equation}
L_{cls} = -\sum_{i=1}^{N} Y_{i}\log \left(P_{i}\right) + L_{sa} + L_{cnce}.
\end{equation}%

Based on the sample distribution of multimodal data, we can flexibly balance each loss function and its combination function on the data within and between samples.

\begin{figure*}[tbp]
	\centering
    \includegraphics[width=\linewidth,height=0.25\linewidth]{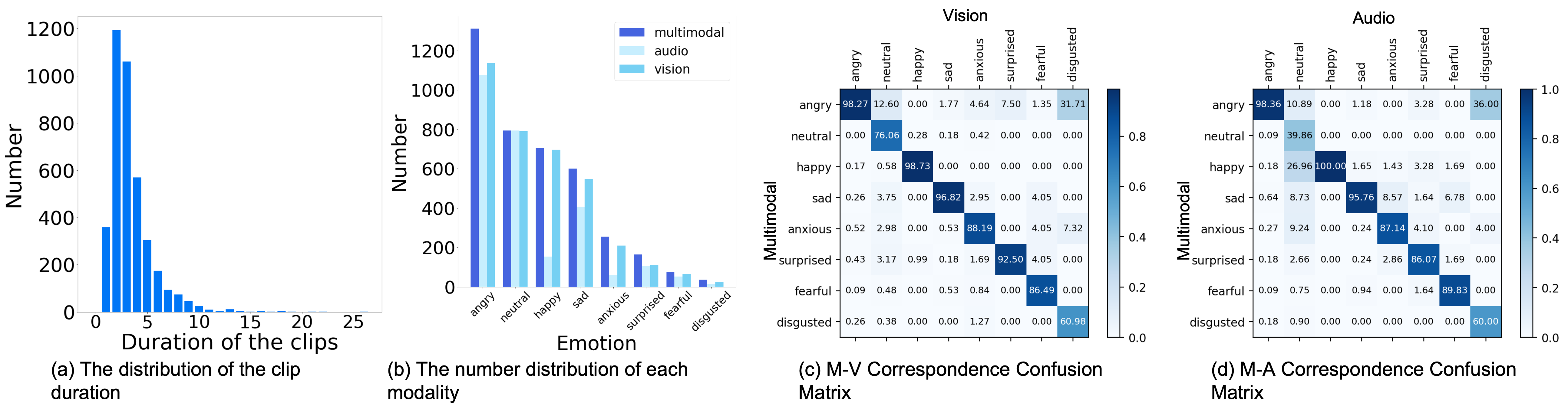}%
	\caption{(a) The temporal distribution of video clips in SDME dataset. (b) The distribution of each category and its sub-modal categories in the dataset. (c) The correlation confusion matrix of multimodal features and visual features in each category. (d) The correlation confusion matrix of multimodal features and audio features in each category.}
	\label{fig2}
	\vspace{-0.2in}
\end{figure*}

\section{4 SDME Dataset}
\subsection{4.1 Problem Definition}
Due to the strong subjectivity and difficulty in collecting and labeling emotional data, most of the existing emotional classification datasets are either continuous emotional states or multi-label. They are usually built from the film reviews, monologues and script performances. As the semantics of each modal emotional expression is similar, it is difficult to reflect the emotional expression in real interactive scenarios. In order to fill the vacancy of a single-label multimodal emotion dataset, we propose a single-label multimodal emotion dataset named SDME to accurately reflect the emotional state in the real interactive scenarios. We collect a large number of high-quality emotional samples from TV dramas that are close to the real interactive scenarios. Moreover, we ensure that each sample contains only a single obvious emotional state. To facilitate the emotional semantic analysis, the unimodal information of each sample is also annotated.

\begin{table}
\centering
\caption{Statistics of SDME Dataset}
\begin{tabular}{lc}
\toprule
Item  & Number \\
\midrule
raw videos       & 300  \\
clips after rough screening       & 66385  \\
clips after manual screening      & 26896  \\
experimental clips    & 3667  \\
Average length of clips (s)   & 3.36  \\
\bottomrule
\vspace{-0.17in}
\end{tabular}
\label{tab:plain1}
\end{table}
\begin{table}
\centering
\caption{Statistics of each category and its sub-modal samples in the SDME dataset for experiments}
\begin{tabular}{cccc}
\toprule
Category  & Multimodal & Vision & Audio\\
\midrule
Neutral       & 795 & 791 & 794 \\
Happy       & 705 & 697 & 153 \\
Sad    & 600 & 548 & 407 \\
Angry   & 1312 & 1137 & 1077 \\
Anxious    & 255 & 209 & 61 \\
\midrule
ALL & 3667 & 3382 & 2492\\
\bottomrule
\end{tabular}
\label{tab:plain2}
\vspace{-0.2in}
\end{table}

\subsection{4.2 Dataset Collection}

To build the SDME dataset, we collect the target clips from various TV series, including romance, action, suspense, fantasy, etc. It guarantees the authenticity and difference of emotion data, which helps the model to capture the relevant features in emotional variations. We got 300 TV episodes, each of which is about 40 minutes. The shot boundary detection model Transnet~\cite{souvcek2020transnet} is used to split the video into short clips, and each episode generates over 800 clips. To obtain the single-label datasets with high quality, we impose the following constraints:
\begin{itemize}
\item Ensure there is only single speaker's voice in each clip.
\item Only the frontal face appears in each clip.
\item The speaker's face in each clip is shown for more than one-third of its length.
\end{itemize}

We grouped the screened 66,385 video clips into eight emotion categories of neutral, happy, sad, angry, anxious, surprised, fearful and disgusted. Moreover, each category was further divided into two intensities to enable better dataset screening and balance in the end. At the same time, we removed the invalid data and the compound emotion data in the labeling process to improve the quality of the single-label dataset.

To avoid the influence of human subjective perception as much as possible, we annotated audio, silent video and multimodal labels separately, which consumed a lot of time. For yielding high quality data, we trained four annotators and performed test annotations in a back-to-back manner. The overall consistency of the final label is 98.50\%, which indicates that the samples are reliable with high quality. 

\subsection{4.3 Dataset Statistics}

The 66,385 video clips have been coarsely screened by the algorithm, which are manually annotated to obtain the valid ones at the later stage. There are 26,896 valid video clips that mainly have the emotions of neutral, happy, sad, angry and anxious. Among them, the neutral emotion category accounts for about 85.4\% of the whole dataset. Moreover, the proportions of other emotions are relatively low, which fully reflects the difficulty in obtaining the single-label emotional data.

Table \ref{tab:plain1} summarizes the basic statistical information. We balance the dataset according to the intensity annotations for various emotions, where 3941 high-confidence samples are selected. Figure \ref{fig2}(a) shows the duration distribution of the selected samples. It can be seen that most of clips are less than 5 seconds. Figure \ref{fig2}(b) shows the quantitative relationship between the number of multimodal category samples and their corresponding sub-modal category samples in the dataset. It fully reflects the universality of the irrelevance of multimodal semantics in the emotional expression of real scenes. To reflect the degree of relevance between modalities more intuitively, we use confusion matrixes in Figure \ref{fig2}(c) and Figure \ref{fig2}(d) to estimate the category-consistency of multimodal and visual, multimodal and audio modality, respectively. Without neutral emotion, only 88.72\% of visual semantics and 59.47\% of audio semantics are consistent with the semantics of multimodal emotion. If all modal information is fully fused, a large number of noise factors will undoubtedly be added, which will greatly affect the performance of the model.

Finally, we selected the five categories of high-confidence samples with the largest number for follow-up experiments, with a total of 3667 samples, as shown in Table \ref{tab:plain2}.

\section{5 Experiments}
\subsection{5.1 Experimental Setups}


\begin{table*}
\centering
\caption{Experimental results on MOSEI, SIMS and SDME datasets for Multimodal emotion Analysis}
\setlength{\tabcolsep}{3mm}{
\begin{tabular}{ccccccc}
\toprule
\multirow{3}{*}{Model} & \multicolumn{2}{c}{MOSEI} & \multicolumn{2}{c}{SIMS} & \multicolumn{2}{c}{SDME} \\
\cmidrule(rr){2-3}
\cmidrule(rr){4-5}
\cmidrule(rr){6-7}
& Acc & F1-Score & Acc & F1-Score & Acc &  F1-Score \\ 
\midrule
TFN & 70.35 ± 0.37 & 69.42 ± 0.64 & 78.77 ± 1.81 & 78.75 ± 1.55 & 65.34 ± 1.21 & 66.20 ± 1.31 \\
LMF & 71.57 ± 0.64 & 70.92 ± 0.81 & 79.26 ± 1.40 & 78.8 ± 1.37 & 65.56 ± 1.45 & 67.64 ± 2.30 \\
EF-LSTM & 70.68 ± 0.55 & 69.82 ± 0.77 & 69.37 ± 0.00 & 56.82 ± 0.00 & 50.41 ± 0.30 & 57.62 ± 0.34 \\
MFN & 70.96 ± 0.58 & 70.14 ± 0.81 & 75.54 ± 2.02 & 75.75 ± 1.67 & 62.56 ± 0.61 & 64.31 ± 0.32 \\
Graph-MFN & 70.25 ± 1.67 & 69.15 ± 2.36 & 76.24 ± 1.74 & 75.93 ± 1.39 & 63.33 ± 1.11 & 65.65 ± 1.55\\
MULT & \textbf{72.16 ± 1.06} & 71.44 ± 1.48 & 77.81 ± 1.89 & 77.39 ± 3.06 & 63.92 ± 1.45 & 66.98 ± 3.23 \\
MISA & 72.02 ± 1.59 & 71.35 ± 2.09 & 76.33 ± 1.72 & 76.31 ± 1.35 & 49.53 ± 1.31 & 52.33 ± 1.65 \\
\midrule
MRE(Ours) & 71.24 ± 1.10 & \textbf{72.15 ± 0.71} & \textbf{81.14 ± 1.19}  & \textbf{81.09 ± 1.44} & \textbf{67.41 ± 1.12}  & \textbf{69.60 ± 1.23} \\
\bottomrule
\end{tabular}
}
\label{tab:plain3}
\vspace{-0.1in}
\end{table*}

\paragraph{Datasets.} For a fair comparison, we employ two commonly used testbeds including MOSEI~\cite{zadeh2018multimodal}, SIMS \cite{yu2020ch}, and our collected SDME dataset in our experiments. MOSEI is a monologue video dataset, where characters make their own comments on a topic. Due to the lack of emotional interaction, the modalities in this dataset are highly relevant, which often have the same semantics. Differently from MOSEI, both SIMS and SDME are collected from the real interactive scenarios. The emotional expression of characters in real interactive scenarios is more complex. They often have large semantic differences among various modalities, which can better reflect the effectiveness of extracting core semantics in our work. We comprehensively evaluate our model using the semantic datasets of specific scenes and irrelevant semantic datasets of the real interactive scenarios.

\paragraph{Networks and Training.} As in~\cite{yu2020ch}, we employ BERT~\cite{devlin2018bert} to extract the 768-dimension text features. Moreover, we use MFCC, zero-crossing rate and Constant-Q chromatogram to form the 33-dimension audio features. OpenFace~\cite{baltrusaitis2018openface} is employed to extract the 35-dimension visual features. We set the learning rate to $5\times 10^{-5}$. The dropout rate is $0.2$. To facilitate the contrastive learning, we set the batch size to $64$. A 3-layer $DNN$ is used to extract visual and audio features. $LSTM$ is employed to learn the textual information. For the extracted features, we suggest a multimodal multi-instance contrastive evaluation network to model the semantics. Finally, we perform rank dimensionality reduction on the semantically weighted features and multimodal fusion. 

\paragraph{Evaluation.} We only use the discrete multimodal annotation for supervision, which treats the emotion analysis as a classification task. To this end, we report the two-class accuracy and F1-Score for MOSEI and SIMS. Moreover, we report the five-class accuracy and F1-Score for SDME. The standard deviation of accuracy is given as a reference. In our experiments, we conduct the tests five times, and take the mean value as the experimental result.

\begin{table}
\centering
\caption{Runtime analysis of MULT, MISA and our method with the same input features on MOSEI}
\begin{tabular}{cc}
\toprule
Model  & Runtime(s) \\
\midrule
MULT       & 2535.73  \\
MISA      & 871.00  \\
MRE(Ours)       & \textbf{95.79}  \\
\bottomrule
\end{tabular}
\label{tab:plain4}
\vspace{-0.1in}
\end{table}

\subsection{5.2 Quantitative Results}

Table \ref{tab:plain3} shows our experimental results. It can be seen that our proposed model achieves the best results in the real interactive scene datasets like SIMS and SDME with large modal semantics irrelevant noises. On the SIMS dataset, the accuracy of our model is 1.88\% higher than the suboptimal model $LMF$ and 2.29\% higher on F1-Score. For the evaluation of the SDME dataset, the accuracy of our presented model is 1.85\% higher than the suboptimal $LMF$ and 1.96\% higher on F1-Score. These results indicate that our model is capable of capturing the relevant semantics under noisy conditions. For the MOSEI dataset with strong relevance between modalities, We obtained the best F1-Score. Based on the analysis of the relevance dataset, we evaluated and compared the runtime of $MULT$ and $MISA$ models, as shown in Table \ref{tab:plain4}. To ensure the fairness, we employ multimodal data with the same feature embedding as input. Our model runs around 26.5 times faster than $MULT$ and 9 times faster than $MISA$. Above all, our proposed approach is able to capture the relevant semantics efficiently under a lightweight conditions, especially in the case of weak modal semantic relevance.

\subsection{5.3 Ablation Study}

To investigate the effectiveness of our proposed approach more specifically, we conduct ablation experiments as shown in Table \ref{tab:plain5}. It can be observed that we can capture relevant semantic information well on SIMS and SDME with weak semantic relevance between modalities when only the $L_{rs}$ loss used. The accuracy is improved by 1.23\% and 0.93\%. Moreover, the F1-Score is improved by 1.94\% and 0.66\%. When only the $L_{cnce}$ loss is used, our model improves the accuracy by 0.4\%, 2.89\% and 0.5\% on MOSEI, SIMS and SDME, respectively. Also, we improve the F1-Score by 0.26\%, 2.85\% and 0.62\%, respectively. When both $L_{rs}$ loss and $L_{cnce}$ loss are employed, our model achieves 0.52\%, 3.42\%, and 1.91\% improvements in accuracy for MOSEI, SIMS, and SDME dataset, and 0.31\%, 2.5\%, and 1.94\% on F1-Score, respectively. 

\begin{table}
\centering
\caption{Ablation study on SA and CNCE loss}
\begin{tabular}{llrr}
\toprule
Datasets & Tasks  & Acc  & F1-Score \\
\midrule
\multirow{4}{*}{MOSEI} 
& Baseline & 70.72 ± 1.35 & 71.75 ± 0.77 \\
& $L_{rs}$  & 70.49 ± 0.60 & 71.64 ± 0.41 \\
& $L_{cnce}$  & 71.12 ± 1.11 & 72.10 ± 0.73 \\
& $L_{rs} + L_{cnce}$ & \textbf{71.24 ± 1.10}  & \textbf{72.15 ± 0.71} \\
\midrule
\multirow{4}{*}{SIMS} 
& Baseline & 77.72 ± 1.63 & 77.59 ± 2.13 \\
& $L_{rs}$ & 78.95 ± 0.09 & 79.53 ± 0.26 \\
& $L_{cnce}$ & 80.61 ± 2.94 & 80.44 ± 3.18 \\
& $L_{rs} + L_{cnce}$ & \textbf{81.14 ± 1.19}  & \textbf{81.09 ± 1.44} \\
\midrule
\multirow{4}{*}{SDME} 
& Baseline & 65.50 ± 0.72 & 67.66 ± 1.11 \\
& $L_{rs}$ & 66.43 ± 1.10 & 68.32 ± 1.66 \\
& $L_{cnce}$ & 66.00 ± 1.01 & 68.28 ± 1.44 \\
& $L_{rs} + L_{cnce}$ & \textbf{67.41 ± 1.12} & \textbf{69.60 ± 1.23} \\
\bottomrule
\end{tabular}
\label{tab:plain5}
\vspace{-0.1in}
\end{table}



\section{6 Conclusion}

In this work, we proposed a multimodal relevance estimation network. To this end, a single-label discrete multimodal emotion dataset SDME was collected for multimodal emotion analysis, where the discrete multi-category samples and rich label attributes provided abundant data for multimodal relevance studies. We suggested a relevant semantic learning scheme to extract semantic information of each sub-modality. Moreover, a contrastive learning method was introduced to bridge the semantic gap of heterogeneous multimodal features in latent feature space. Extensive experimental evaluations demonstrated the efficacy of our proposed approach, which is capable of effectively capturing the modality-relevant semantic information, especially in the case of large modality class bias. 

Since the temporal information was neglected, our presented method is not significant enough in the strong relevant semantic scenarios. In future work, we will make use of temporal features to further extract multimodal relevant semantic information.

\newpage
\bibliography{aaai23}

\end{document}